\pdfoutput=1

\documentclass[11pt]{article}

\usepackage[preprint]{acl}

\usepackage{times}
\usepackage{latexsym}

\usepackage[T1]{fontenc}

\usepackage[utf8]{inputenc}

\usepackage{microtype}

\usepackage{inconsolata}

\usepackage{graphicx}

\usepackage{hyperref}
\usepackage{multirow}
\usepackage{tabularx}
\usepackage{makecell}
\usepackage{colortbl}
\usepackage{amsmath}
\usepackage{amssymb}
\usepackage{mathtools}
\usepackage{amsthm}
\usepackage{microtype}
\usepackage{subfigure}
\usepackage{booktabs} 
\usepackage{xspace}

\usepackage{fancyhdr}
\usepackage{xcolor}
\usepackage{algorithm}
\usepackage{algorithmic}
\usepackage{natbib}
\usepackage{eso-pic}
\usepackage{forloop}
\usepackage{url}

\newcommand{\hide}[1]{}
\newcommand{\vpara}[1]{\vspace{1.5ex}\noindent\textbf{#1}}
\newcommand{\model}{SAM-Decoding\xspace}
\newcommand{\smodel}{SAM-Decoding }

\title{SAM Decoding: Speculative Decoding via Suffix Automaton}

\author{
    Yuxuan Hu\textsuperscript{\rm 1,\rm 2},\ Ke Wang\textsuperscript{\rm 1,\rm 2},\ Xiaokang Zhang\textsuperscript{\rm 1,\rm 2},\ Fanjin Zhang\textsuperscript{\rm 4}\\
    {\bf Cuiping Li}\textsuperscript{\rm 1,\rm 3},\ {\bf Hong Chen}\textsuperscript{\rm 1,\rm 3},\ {\bf Jing Zhang}\textsuperscript{\rm 1,\rm 3}\thanks{Corresponding author. zhang-jing@ruc.edu.cn} \\
    \textsuperscript{\rm 1}School of Information, Renmin University of China, Beijing, China \\
    \textsuperscript{\rm 2}Key Laboratory of Data Engineering and Knowledge Engineering, Beijing, China \\
    \textsuperscript{\rm 3}Engineering Research Center of Database and Business Intelligence, Beijing, China \\
    \textsuperscript{\rm 4}Knowledge Engineering Group, Tsinghua University, Beijing, China \\
}

\begin{document}
\maketitle
\begin{abstract}
Speculative decoding (SD) has been demonstrated as an effective technique for lossless LLM inference acceleration.
Retrieval-based SD methods, one kind of model-free method,
have yielded promising speedup,
but they often rely on incomplete retrieval resources, inefficient retrieval methods, and are constrained to certain domains.
This paper presents a novel retrieval-based speculative decoding method that adapts suffix automaton (SAM) for efficient and accurate draft generation by utilizing common text corpus and dynamic text sequence.
Unlike existing $n$-gram matching methods, SAM-Decoding finds the exact longest suffix match, achieving an average time complexity of O(1) per generation step of SAM update and suffix retrieval.
It can also integrate with existing methods, adaptively selecting a draft generation strategy based on match length to generalize to broader domains.
Extensive experiments on Spec-Bench show that our method is $18\%+$ faster than other retrieval-based SD methods. Additionally, when combined with advanced EAGLE-2, it provides an additional speedup of $3.28\%$ -- $11.13\%$ across various-sized LLM backbones. Our code is available at our \href{https://github.com/hyx1999/SAM-Decoding}{repository}.

\end{abstract}

\section{Introduction}

The Transformer-based Large Language Models (LLMs)~\cite{gpt3-brown-2020, llama3-dubey-2024, qwen2-yang-2024} have demonstrated remarkable abilities and are extensively adopted in numerous domains.
The scaling law drives LLMs to become deeper, reaching hundreds of billions of parameters, which makes them inefficient for generating text in a token-by-token autoregressive manner.
Speculative decoding methods~\cite{spec-decoding-leviathan-2023,medusa-cai-2024} seek to tackle this problem by quickly generating multiple draft tokens and subsequently concurrently verifying them with LLMs. 
These methods can decrease inference latency substantially while maintaining decoding accuracy.

Speculative methods can be categorized into model-based and model-free methods.
Model-based methods need to carefully choose and train one or more small-sized draft models.
For example, Medusa~\cite{medusa-cai-2024} utilizes multiple decoding heads to generate multiple future tokens while EAGLE-2~\cite{eagle2-li-2024} leverages shallow Transformer layers to predict the next last hidden states and corresponding decoding tokens.
Although these methods achieve impressive speedup, 
they often fail to generate long draft tokens due to drafting overhead or decaying prediction accuracy.
Retrieval-based speculative decoding methods,
a major type of model-free methods,
aim to remedy this issue by generating draft tokens from text corpus or current text sequence.


\begin{figure}[t]
    \centering
    \includegraphics[width=\linewidth]{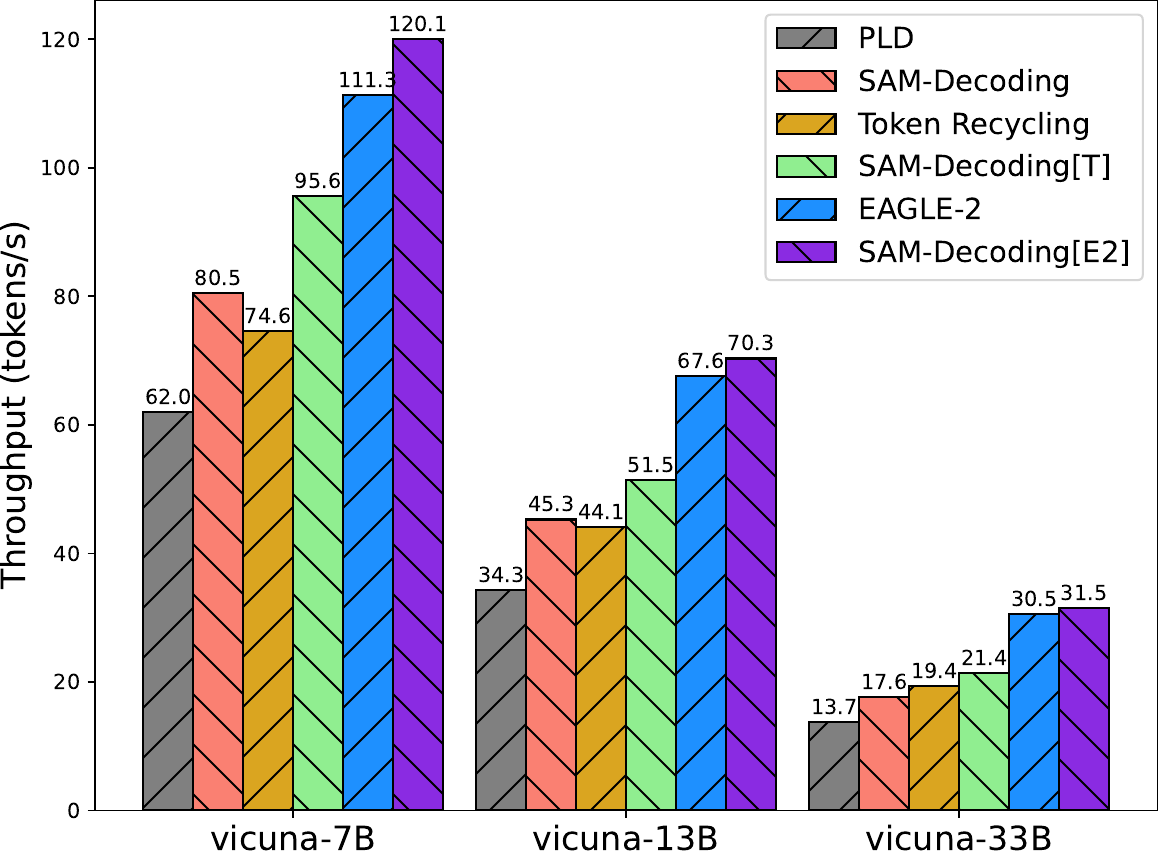} 
    \caption{Throughput of Vicuna-7B, Vicuna-13B, Vicuna-33B on MT-Bench with A6000 GPU using PLD, Token  Recycling~\cite{token-recycle-luo-2024}, EAGLE-2, and SAM-Decoding, where PLD is the SOTA retrieval-based SD baseline.}
    \label{fig: throughput}
\end{figure}


However, current retrieval-based methods have notable limitations. 
\textbf{Firstly}, diverse retrieval sources contribute to the efficiency of retrieval-based SD methods, but existing methods typically rely on a single retrieval source: PLD~\cite{pld-saxena-2023} focuses on current text while REST~\cite{rest-he-2024} uses a text corpus.
\textbf{Secondly}, the retrieval techniques they use have efficiency limitations.
PLD finds $n$-gram matching from current text sequence, 
but it has poor theoretical computational complexity and limited applicability to larger text corpus.
REST uses suffixed arrays, which provides better complexity than PLD, but still not optimal complexity.
\textbf{Thirdly}, retrieval-based methods are suitable for specialized domains (e.g., summarization and RAG), which are unable to bring a noticeable acceleration in other domains.


To address limitations in previous retrieval-based methods, this paper introduces \model, an innovative speculative decoding technique based on suffix automaton.
(1) To enhance the coverage of the retrieved corpus,
we utilize the common text corpus and the current text sequence as retrieved sources.
(2) To improve the retrieval efficiency and accuracy,
we adapt a suffix automaton (SAM) to solve the longest suffix match problem,
which yields more accurate match positions and exact match length compared to $n$-gram matching.
As for retrieval efficiency,
the average time complexity of SAM update and suffix retrieval is $O(1)$ by capturing relationships between adjacent suffixes.
(3) To generalize our method,
assuming the matching length of the longest suffix implying the quality of retrieval draft tokens,
our method can be integrated with other types of speculative decoding methods,
enabling more efficient text generation by deciding whether to adopt auxiliary decoding techniques.

Specifically, \model creates both a static suffix automaton for the text corpus and a dynamic suffix automaton for the current text sequence.
The nodes of suffix automaton represent substrings in the text corpus or current sequence.
The earliest position of each substring is recorded in each node.
During generation, we can directly retrieve and filter drafts from the context using the matching positions and longest suffixes' matching length.
After each generation step, the automaton is updated: static automaton nodes transition based on new tokens, 
while the dynamic automaton first expands its structure before node transitions.



Extensive evaluations demonstrate the competitive performance of our method across tasks.
On Spec-Bench, \model achieves $18\%+$ faster than previous retrieval-based speculative decoding methods (e.g., PLD, REST, etc.).
\model further achieves speedups of up to $1.3\times$ over alternative baselines on the code-generation benchmark like HumanEval.
When combined with EAGLE-2 \cite{eagle2-li-2024}, as shown in Figure~\ref{fig: throughput}, our method outperforms the state-of-the-art, delivering an additional $3.28\%$ -- $11.13\%$ speedup on MT-Bench w.r.t. various LLM backbones.


\section{Background}

\begin{figure}[t]
    \centering
    \includegraphics[width=\linewidth]{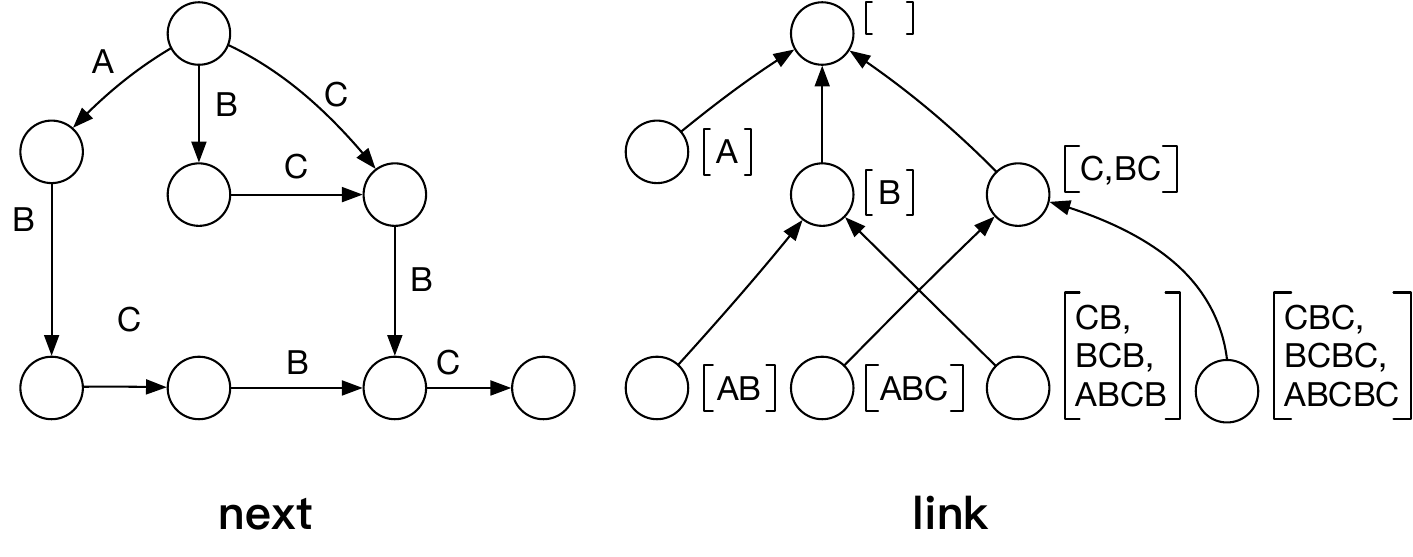} 
    \caption{The suffix automaton corresponding to the string ``ABCBC''.}
    \label{fig: example suffix automation}
\end{figure}

\subsection{Suffix Automaton}

Suffix Automaton is an efficient data structure for representing the substring index of a given string, 
which allows fast substring retrieval.
The time complexity of constructing a suffix automaton is $O(L)$, where $L$ is the length of the string and it can be constructed incrementally.

As shown in Figure \ref{fig: example suffix automation}, a suffix automaton contains a series of nodes and two types of state transfer edges, \textbf{extension edges (next)} and \textbf{suffix link edges (link)}.
A node in the automaton represents a state and corresponds to all substrings that have the same ending position in the string. Meanwhile, 
extension edges are standard edges that represent a possible extension of the current substring by appending a new character,
while suffix link edges create a path that allows the automaton to quickly jump to states representing shorter suffixes of the current substring.

Based on the two types of transfer edges, for a progressively generated token sequence, we can find the longest suffix that matches the sequence in a suffix automaton at each step of the generation with an average $O(1)$ time complexity.

\begin{figure*}[t]
    \centering
    \includegraphics[width=0.9\textwidth]{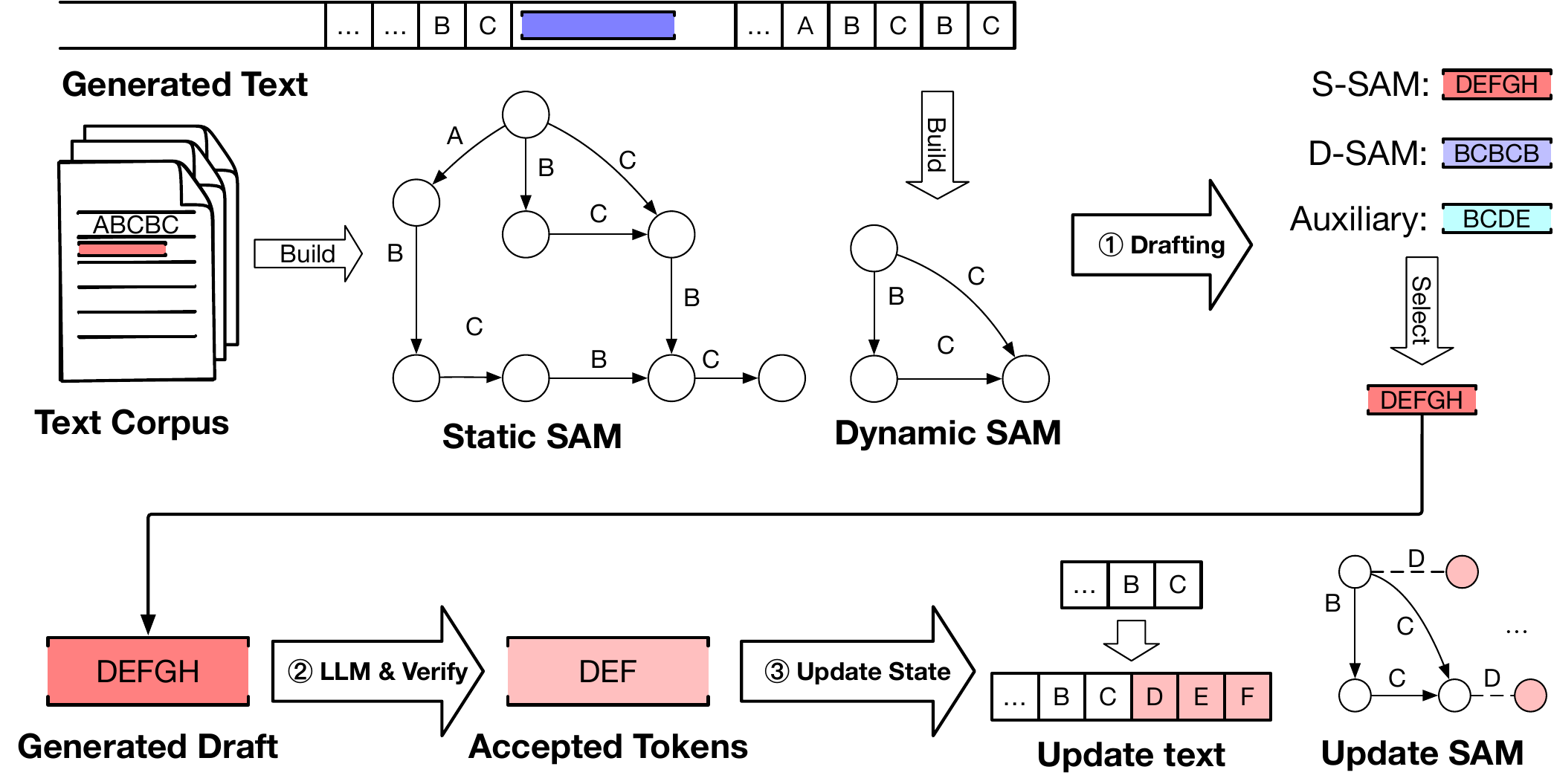} 
    \caption{Overview of \model's workflow. In each round of generation, the suffix automaton matches the suffixes of the generating text and retrieves the draft from the text corpus and the generated text respectively according to the matching position.
    Our method can be combined
    with an auxiliary SD algorithm (Auxiliary) to deal with the scenarios where the retrieval is not applicable. We select the best draft from the three candidate drafts based on the match length, and then the drafts are verified by the LLM for accepted tokens. Using these accepted tokens, we finally extend the dynamic SAM and generate text for the next round of generation. 
    }.
    \label{fig: SAM-Decoding}
\end{figure*}

\subsection{Speculative Decoding}

Given the model input \( x = (x_1, x_2, \ldots, x_t) \), an LLM generates a new token $x_{t+1}$ at each generation step autoregressively. 
\hide{
\begin{equation}
    x_{t+1} = \text{Sample}(\text{LLM}(x)[t]).
\end{equation}
}
The key idea of speculative decoding is to utilize a lightweight draft model to generate multiple candidate tokens quickly, i.e., 
\( x_{\text{draft}} = (x_{t+1}, x_{t+2}, \ldots, x_{t+n}) \),
and then the target LLM simultaneously evaluates these candidates
and accept those aligned with the output distribution of the LLM, i.e., \( x_{\text{accept}} = (x_{t+1}, x_{t+2}, \ldots, x_{t+m}) \), where $n$ and $m$ denote the size of the draft and the number of accepted tokens.

\hide{
containing multiple tokens based on the draft model and the input \( x \). The drafts are then verified by the LLM, and the correct tokens \( x_{\text{accept}} = (x_{t+1}, x_{t+2}, \ldots, x_{t+m}) \) are accepted:
\begin{align*}
    x_{draft} & = \text{Draft}(x), \\
    x_{accept} & = \text{Verify}(\text{LLM}(x, x_{draft})[t:t+m]).
    \label{Eq. verify}
\end{align*}
}

\hide{
By checking whether each token in the draft is the predicted result of the previous tokens in the LLM, we can filter out the accepted tokens. Specifically, we find the maximum number of consecutive tokens \( m_{\text{accept}} \) such that:
\begin{align*}
    & \text{Sample}(\text{LLM}(x, x_{draft})[t+i-1]) = x_{draft}[i] \\
    & \quad i \in \{1, 2, ..., m_{accept}\}.
\end{align*}
}


In the above, we assume that the draft is a sequence of tokens. Recent works proposed to verify a candidate token tree via a tree mask in the attention module to make the target LLM simultaneously evaluate multiple branches of this token tree,
thereby increasing the acceptance length of the draft model.

\section{\model}

In this section, we introduce our proposed method, \model. \smodel is a retrieval-based speculative decoding method designed to address three key limitations in existing retrieval-based speculative methods: \textbf{(1)} The use of 
insufficient
retrieval sources. \textbf{(2)} The employment of inefficient retrieval methods and restrictions on $n$-gram matching lengths. \textbf{(3)} Subpar performance outside specialized domains (e.g., summarization and RAG tasks).

To tackle the first two limitations, \smodel leverages suffix automaton on diverse text sources, which significantly enhances the 
coverage of retrieved corpus
and the efficiency of the retrieval process while allowing for flexible matching lengths. In what follows, we detail how \smodel can be integrated with both model-free and model-based methods. By utilizing the precise matching information provided by the suffix automaton, our method not only overcomes the third limitation but also ensures consistent performance improvements across a wide range of tasks. The workflow of SAM-Decoding is shown in Figure~\ref{fig: SAM-Decoding}.

\subsection{Suffix Automaton Construction}


To cover comprehensive retrieval sources,
SAM-Decoding builds suffix automaton (SAM) by utilizing common text corpus and the current text sequence (including user prompts and already generated tokens).
Thus, we construct two types of suffix automaton: a \textbf{static suffix automaton} and a \textbf{dynamic suffix automaton}. 
For the text corpus, we pre-build a static suffix automaton offline, which is used for state matching during inference. 
For the current text sequence, we create and expand a dynamic suffix automaton incrementally as generation progresses and performing state matching concurrently.

A suffix automaton can be constructed in linear time using Blumer's algorithm~\cite{sam-blumer-1984}. Since the suffix automaton is designed for a single reference string, static suffix automation can not be directly built using a text corpus.
To this end, we concatenate multiple strings in the corpus by using special symbols like an End-of-Sentence (EOS) token.
We then construct a static suffix automaton for this concatenated string.

We have modified the suffix automaton for better draft generation. 
During each generation step, the current generated text sequence corresponds to a node in the automaton, representing the longest suffix match.
At each node of the suffix automaton, we record the earliest position of all substrings corresponding to that node in the reference string, termed as \textbf{min\_endpos},
which allows us to efficiently locate the previous ending position of the matched longest suffix.
Hereafter, the subsequent tokens after the matched suffix can be regarded as potential drafts.
The construction process of the suffix automaton is detailed in Appendix \ref{appendix: Construction Process of Suffix Automaton}.

For the static suffix automaton, based on the frequency of occurrence of different substrings, we additionally compute the top-k successor states (\textbf{topk\_succ}) of each state, and subsequently use them to construct more complex tree drafts. 
Although computing the successor states requires significant computation, this can be done offline, eliminating the need to account for this time overhead in real-time processing.


\subsection{Drafting with Suffix Automaton}




We illustrate how to generate draft tokens efficiently based on the built suffix automaton.
Let \( S \) denote the suffix automaton, \( T \) denote its associated reference text,  and  \( x = (x_1, x_2, \ldots, x_t) \) denote the current text sequence.
The state within the suffix automaton corresponding to the sequence \( x \) is denoted as \( s_t \). In each round of generation, the transition to the next state is performed based on the newly generated token \( x_{t+1} \) and the current state \( s_t \):
\begin{align*}
    s_{t+1} & = \text{Transfer}(S, x_{t+1}, s_t).
\end{align*}

For dynamic suffix automaton, we extract \( n \) consecutive tokens from the reference text \( T \) to form a draft, using the \textbf{min\_endpos} value stored in the node corresponding to state \( s_{t+1} \), termed as \( p_{t+1} \). Then the draft \( d_{t+1} \) is defined as:
\begin{align*}
    d_{t+1} = T[p_{t+1} + 1 : p_{t+1} + n],
\end{align*}
where \( d_{t+1} \) represents the generated draft and \( n \) denotes the length of the draft.

For static suffix automaton, we construct a tree-structured draft by Prim's algorithm based on top-k successors, as detailed in Appendix~\ref{appendix: Drafting via Prim's Algorithm},
\begin{align*}
    d_{t+1} = \text{Prim}(S, s_{t+1}, x_t).
\end{align*}

In practical use, 
we track the longest-matched suffix length (denoted as $l$) to determine whether to use the static suffix automaton or the dynamic suffix automaton. 
Specifically, let \( l_1 \) and \( l_2 \) be the matching lengths of the static and dynamic automata, respectively. 
Our experimental findings indicate that drafts generated from the dynamic automaton often outperform those from the static text corpus. Consequently, we prioritize drafts from the dynamic automaton. 
We use the draft from the static automaton only if \( l_1 > l_2 + l_{\text{bias}} \), where \( l_{\text{bias}} \) is a predefined constant.

The complete state transfer process of the suffix automaton is shown in Algorithm \ref{algo: State Transfer of Suffix Automaton}. 
Using amortized analysis, we can prove that the average complexity of state transfer is \( O(1) \), 
with a worst-case time complexity of
\( O(L) \), where \( L \) is the length of the current generated text (C.f. proof in Appendix \ref{appendix: Time Complexity of Suffix Automaton}). 
Existing methods like
PLD uses a brute-force search for $n$-gram matches,
resulting in a time complexity of \( O(n^2 L) \). 
REST also employs n-grams but searches using suffix arrays, 
leading to a time complexity of \( O(n^2 \log L) \). Here, \( n \) is the predefined maximum matching length, and \( L \) is the length of the current text 
or the concatenated texts in the corpus.
In contrast, our proposed \model  model has a lower time complexity and can find the exact longest suffix match without any limit on matching length, making it faster and more accurate for draft generation.
\begin{algorithm}[t]
    \caption{State Transfer of Suffix Automaton}
    \label{algo: State Transfer of Suffix Automaton}
\begin{algorithmic}
    \FUNCTION{Transfer}
        \STATE {\bfseries Input:} suffix automaton $S$, next token $t$, current state $s$, current matching length $l$
        \WHILE{$s \neq S.\text{root}$ \AND $t \notin s.\text{next}$}
            \STATE $s,\,l = s.\text{link},\,s.\text{link}.\text{length}$
        \ENDWHILE
        \IF{$t \in s.\text{next}$}
            \STATE $s,\,l = s.\text{next}[t],\,l + 1$
        \ELSE
            \STATE $l = 0$        
        \ENDIF
        \STATE {\bfseries Output:} next state $s$, next matching length $l$
    \ENDFUNCTION
\end{algorithmic}
\end{algorithm}

\subsection{Update of Suffix Automaton}




After the draft is generated, we verify it using the large language model (LLM) and accept the correct tokens, denoted as \( x_{\text{accept}} = (x_{t+1}, x_{t+2}, \ldots, x_{t+m}) \). 
We then update the state of the suffix automaton based on these accepted tokens. For the static suffix automaton, we simply transfer the states according to Algorithm \ref{algo: State Transfer of Suffix Automaton}:
\begin{align*}
    s_{t+i} = \text{Transfer}(S, s_{t+i-1}, x_{t+i}), \; i \in \{1, 2, ..., m\}.
\end{align*}
For the dynamic suffix automaton, we first transfer the matching state based on the accepted tokens and then expand the state. Let \( S_t \) denote the dynamic suffix automaton for the generated text
\( (x_1, x_2, \ldots, x_t) \). The process is as follows:
\begin{align*}
    s_{t+i} & = \text{Transfer}(S_{t+i-1}, s_{t+i-1}, x_{t+i}), \\
    S_{t+i} & = \text{Expand}(S_{t+i-1}, x_{t+i}), \\
    i & \in\{1, 2, ..., m\},
\end{align*}
where the process of expanding the suffix automaton is detailed in Appendix \ref{appendix: Construction Process of Suffix Automaton}.

\subsection{Adaptive Draft Selection}

The retrieval-based speculative decoding methods excel at generating drafts from the corpus or the current text sequence effectively. 
If it fails to produce a satisfactory draft, other speculative decoding techniques can be employed to generate more diverse drafts.
To combine different types of drafts,
a straightforward idea is that the length of the suffix match can indicate 
the confidence
of the draft produced by the automaton,
where long matches imply that more tokens are likely to be acceptable.

To implement this, we concurrently use an auxiliary speculative decoding technique alongside the suffix automaton. 
During each generation step, we adaptively select the drafts offered by the automaton or the auxiliary SD method based on the match length of the generated text within the automaton. For the auxiliary SD method, we set a fixed virtual match length $l_{\rm threshold}$.
In our study, we consider two auxiliary cutting-edge speculative decoding methods:
the model-free Token Recycling and the model-based EAGLE-2. 

Among them, Token Recycling maintains an adjacency list of the top-$k$ probable next tokens for each token
and builds a draft tree using breadth-first search, and it continuously updates the list based on the latest tokens.
EAGLE-2, on the other hand, 
leverages a Transformer decoder layer to jointly predict the last hidden states of the LLM and the next token autoregressively.


\section{Experiments}

\begin{table*}[t]
\newcolumntype{?}{!{\vrule width 1pt}}
\newcolumntype{C}{>{\centering\arraybackslash}p{2em}}
\centering
\renewcommand\arraystretch{1.5}
\resizebox{0.9\linewidth}{!}{
\newcommand{\Bf}[1]{\textbf{#1}}
\begin{tabular}{l|ccc|ccc|ccc}
\hline
\multirow{2}{*}{Method}    & \multicolumn{3}{c|}{Spec-Bench} & \multicolumn{3}{c|}{HumanEval}  & \multicolumn{3}{c}{HAGRID}      \\ \cline{2-10} 
                           & \#MAT & Tokens/s & Speedup      & \#MAT & Tokens/s & Speedup      & \#MAT & Tokens/s & Speedup      \\ \hline
Lookahead*                 & 1.63  & 44.37    & 1.20$\times$ & 1.76  & 30.81    & 1.54$\times$ & 1.46  & 23.58    & 1.32$\times$ \\
REST*                      & 1.63  & 51.34    & 1.38$\times$ & 1.85  & 34.60    & 1.74$\times$ & 1.53  & 24.91    & 1.39$\times$ \\
PIA                        & 2.08  & 55.45    & 1.47$\times$ & 2.62  & 65.49    & 1.68$\times$ & 2.43  & 66.65    & 1.95$\times$ \\
PLD                        & 1.75  & 59.02    & 1.56$\times$ & 1.65  & 59.04    & 1.52$\times$ & 2.03  & 44.11    & 1.29$\times$ \\
\textbf{SAM-Decoding}
                           & \Bf{2.30}  & \Bf{69.37} & \Bf{1.84}$\times$ & \Bf{2.64}  & \Bf{88.91} & \Bf{2.29}$\times$ & \Bf{2.44}  & \Bf{76.72}    & \Bf{2.24}$\times$ \\ \cline{1-10} 
Token Recycling            & 2.83  & 69.65    & 1.84$\times$ & 2.78  & 75.44    & 1.94$\times$ & 2.88  & 66.17    & 1.93$\times$ \\
\textbf{SAM-Decoding{[}T{]}}  
                           & \Bf{3.03}  & \Bf{85.73}  & \Bf{2.27}$\times$ & \Bf{2.94}  & \Bf{95.08}    & \Bf{2.45}$\times$ & \Bf{3.23}  & \Bf{87.93}    & \Bf{2.57}$\times$ \\  \cline{1-10} 
EAGLE-2                    & 4.36  & 90.14    & 2.38$\times$ & \Bf{5.13} & 125.77   & 3.24$\times$ & 4.15  & 82.61    & 2.41$\times$ \\
\textbf{SAM-Decoding{[}E2{]}}
                           & \Bf{4.62}  & \Bf{97.56} & \Bf{2.58}$\times$ & 4.95  & \Bf{130.28} & \Bf{3.35}$\times$ & \Bf{4.75}  & \Bf{96.60}    & \Bf{2.81}$\times$ \\ \hline
\end{tabular}
}
\caption{Inference efficiency of SAM-Decoding compared to the baselines on Spec-Bench, HumanEval, and HAGRID, where * indicates that the method was compared with the baseline provided in its environment.}
\label{table: spec-bench, humaneval, hagrid}
\end{table*}

\begin{figure*}
    \centering
    \begin{minipage}[t]{0.4\textwidth}
        \includegraphics[width=\linewidth]{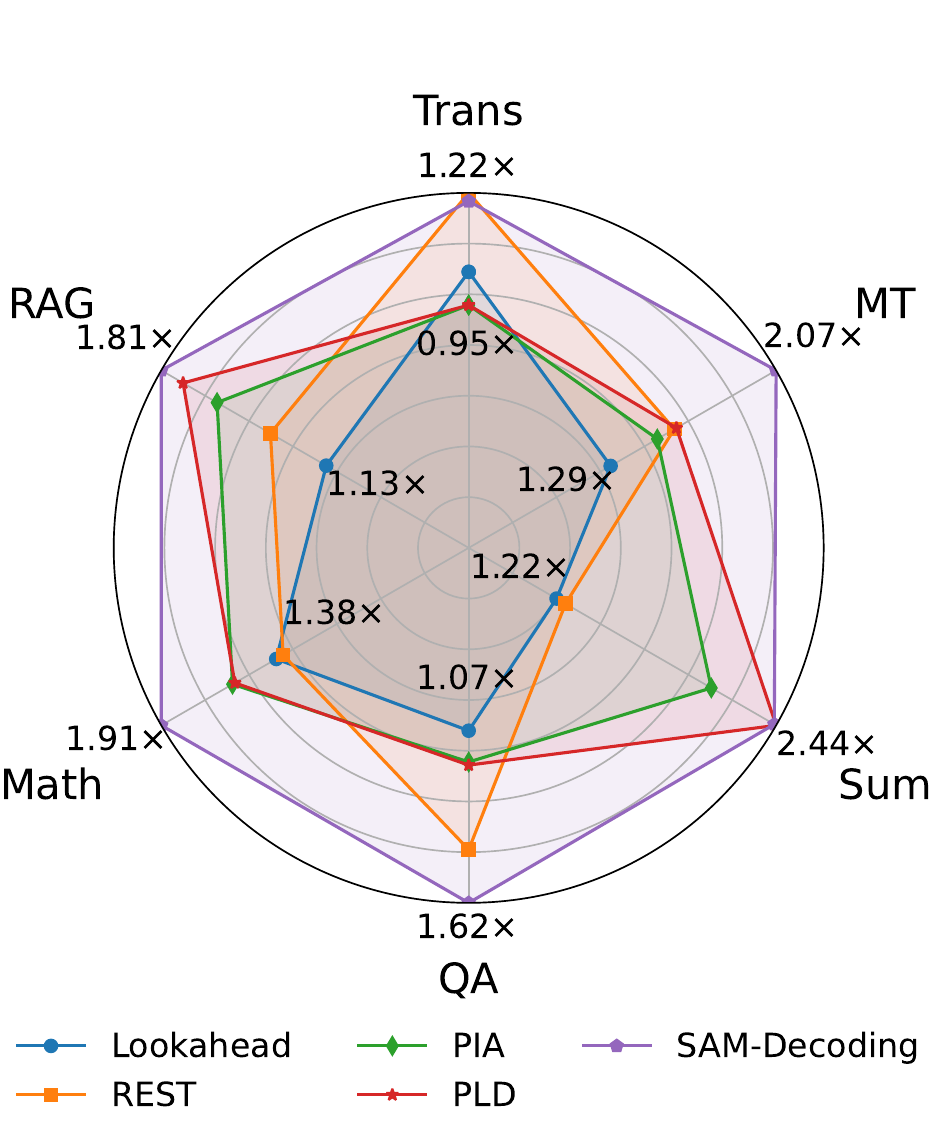} 
        \caption{Relative speedup of SAM-Decoding compared to retrieval-based SD baselines on Spec-Bench.}
        \label{fig: radar retrieval}
    \end{minipage}
    \hfill
    \begin{minipage}[t]{0.5\textwidth}
    \includegraphics[width=\linewidth]{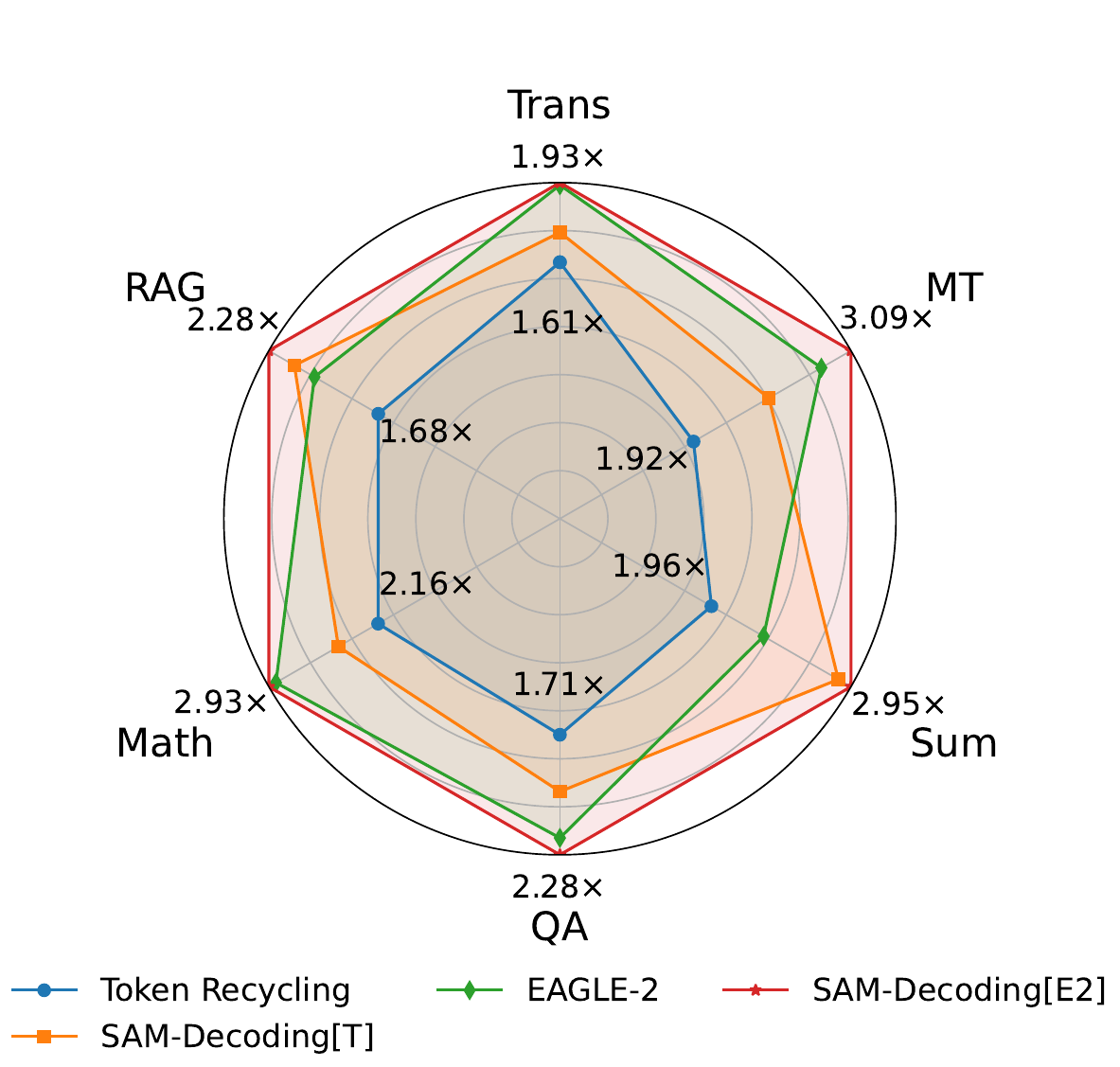} 
    \caption{Relative speedup of SAM-Decoding compared to SD baselines on Spec-Bench when combined with auxiliary SD methods.}
    \label{fig: radar mix}
    \end{minipage}
\end{figure*}

In this section, we first introduce our experimental setup, then present the experimental results, and finally present the ablation experiments.

\vpara{Models and Tasks.} We conducted experiments on Vicuna-7B-v1.3~\cite{mt-bench-zheng-2023}. We evaluated \smodel on Spec-Bench~\cite{spec-bench-xia-2024}, HumanEval~\cite{humaneval-chen-2021}, and HARGID~\cite{hagrid-kamalloo-2023}. Spec-Bench is a comprehensive benchmark designed for assessing Speculative Decoding methods across diverse scenarios. It is based on six commonly used datasets, MT-Bench~\cite{mt-bench-zheng-2023}, WMT14 DE-EN, CNN/Daily Mail~\cite{cnn-daily-nallapati-2016}, Natural Question~\cite{NQ-kwiatkowski-etal-2019}, GSM8K~\cite{gsm8k-cobbe-2021}, and DPR~\cite{DPR-karpukhin-2020}, including six aspects: Multi-turn Conversation (MT), Translation (Trans), Summarization (Sum), Question Answering (QA), Mathmatical Reasoning (Math), and Retrieval-augmented Generation (RAG). In addition, HumanEval, and HARGID are used to evaluate the speed of decoding methods in Code Generation task and Context Q\&A task, respectively.

\vpara{Baselines.} We considered the following baseline methods, including the model-based method EAGLE-2~\cite{eagle2-li-2024}, the model-free method Token Recycling~\cite{token-recycle-luo-2024}, and the retrieval-based methods Lookahead Decoding~\cite{lookahead-fu-2024}, PIA~\cite{pia-lookahead-zhao-2024}, PLD~\cite{pld-saxena-2023} and REST~\cite{rest-he-2024}.

\vpara{Metrics.} We evaluated speculative decoding methods from the following aspects~\cite{eagle-li-2024}
\begin{itemize}
    \item \textbf{Speedup Ratio}: The wall-time speedup ratio of speculative decoding methods compared to autoregressive generation methods.
    \item \textbf{Mean Accepted Tokens}: The average number of tokens accepted per generation step.
    \item \textbf{Throughput}: The average number of tokens generated per second.
\end{itemize}
\vpara{Experiment Setup.} We conducted experiments on a server equipped with a 20-core CPU and a single NVIDIA RTX A6000 GPU (48GB). The experiments were implemented using PyTorch 2.3.0, Transformers 4.46.1 and CUDA 12.1. For the models, we used the float16 data type and applied greedy decoding with a batch size of 1. Regarding hyperparameters, $l_{\rm bias}$ and $l_{\rm threshold}$ were set to 5, but when there is no auxiliary method $l_{\rm bias}$ is set to 0. The size of the draft generated by the automaton was set to 40 by default, while for code datasets the size of the draft is set to 16. For the auxiliary speculative decoding methods, we used the default configurations as described in their respective original papers. 



For \model, we constructed a static suffix automaton based on the Vicuna-7B generation results on datasets Stanford-alpaca, python-code-instruction-18k, and GSK8k. To enhance our model, we incorporated two auxiliary approaches: the model-free Token Recycling and the model-based EAGLE-2. Here, SAM-Decoding[T], and SAM-Decoding[E2] denote the combinations of our base model with Token Recycling, and EAGLE-2, respectively. 

\vpara{Experiment Results.} Experimental results on Spec-Bench, HumanEval and HAGRID when using Vicuna-7B-v1.3 are shown in Table \ref{table: spec-bench, humaneval, hagrid}. It can be seen that \smodel has higher inference speedups on all datasets compared to retrieval-based baselines, achieving speedup ratios of $1.84\times$, $2.29\times$, and $2.24\times$ on each of the three datasets. Meanwhile, further speedups can be achieved by combining \smodel with other types of methods. On the Spec-Bench and HAGRID dataset, the inference speed of Token Recycling and EAGLE-2 can be further improved by combining SAM-Decoding. In Spec-Bench, the speedup ratios are improved from $1.84\times $, $2.38\times$ to $2.27\times$, $2.58\times$, respectively, whereas on HAGRID dataset, the speedup ratios are improved from $1.93\times$, $2.41\times$ to $2.57\times$, $2.81\times$. In the HumanEval dataset, the throughput of the model-based EAGLE-2 method changed slightly after integrating SAM-Decoding, due to the fact that the code generation task is less likely to copy the generated text during the generation process. Fortunately, \smodel can still speedup the model-free method Token Recycling, increasing its speedup ratios from $1.94\times$ to $2.45\times$.

In Figures \ref{fig: radar retrieval} and \ref{fig: radar mix}, we further show the speedup of the different methods on each task of Spec-Bench. Compared to retrieval-based SD baselines, SAM-Decoding shows better performance across all tasks. Meanwhile, in the Spec-Bench, Multi-turn Conversation, Summarization, and Retrieval-augmented Generation were identified as particularly amenable to retrieval techniques. The results indicate that integrating SAM-Decoding into existing method led to notable speed improvements. Specifically, for Token Recycling, the speedup ratio for the three tasks raised from $1.92\times$, $1.96\times$, and $1.68\times$ to $2.48\times$, $2.86\times$, and $2.14\times$, respectively. For EAGLE-2, the speedup ratios raised from $2.87\times$, $2.33\times$, and $2.03\times$ to $3.02\times$, $2.76\times$, and $2.23\times$, respectively.

In addition to Vicuna-7B, we also conducted experiments on more models. Figure~\ref{fig: throughput} shows the throughput of Vicuna-7B, Vicuna-13B and Vicuna-33B on MT-bench using SAM-Decoding and other baseline SD methods. It can be seen that SAM-Decoding outperforms retrieval-based baselines on all models. Also, SAM-Decoding can further improve the inference speed of model-free and model-based SD methods by combining them with SAM-Decoding. For more experimental results, please refer to Appendix \ref{section: Additional Experiment Results}.

\begin{figure}[t]
    \centering
    \includegraphics[width=0.8\linewidth]{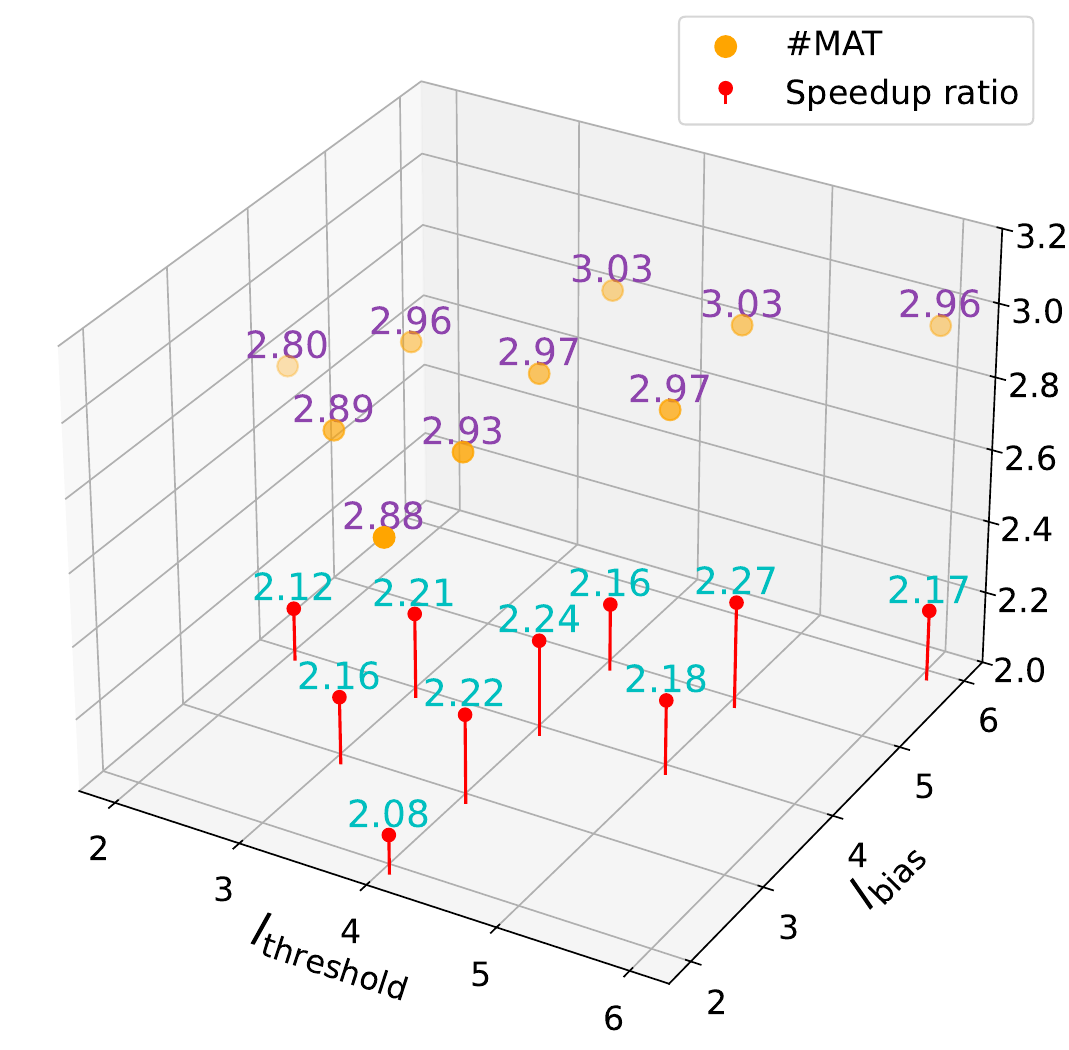} 
    \caption{The speedup ratio and mean accepted toknes of SAM-Decoding[T] under different $l_{\rm bias}$ and $l_{\rm threshold}$.}
    \label{fig: SAM-Decoding bias and threshold}
\end{figure}

\begin{figure}[t]
    \centering
    \includegraphics[width=0.8\linewidth]{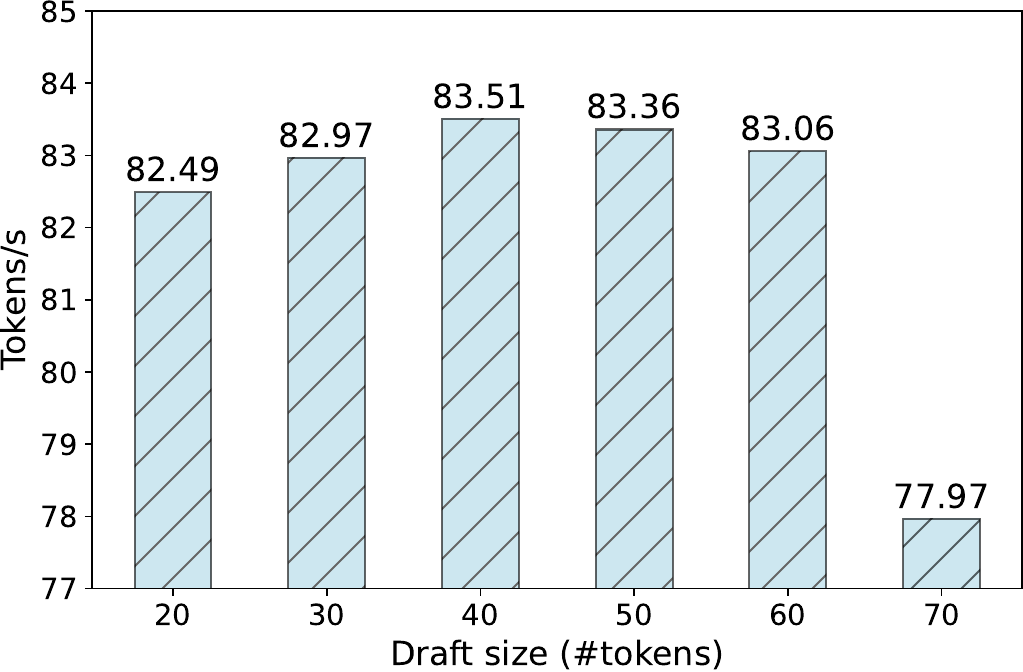}
    \caption{The throughput of SAM-Decoding[T] under different draft size.}
    \label{fig: SAM-Decoding draft size}
\end{figure}

\vpara{Ablation Experiments.} To further understand the contributions of various components of \smodel and the influence of different hyperparameters on inference speed, we conducted a series of ablation studies. 


Firstly, we examined the effects of $l_{\rm bias}$ and $l_{\rm threshold}$ on inference speed through a grid search. These parameters control the preference for generating draft from the current text over text corpus and the preference for using suffix automaton over the auxiliary SD method when creating drafts. The findings are summarized in Figure \ref{fig: SAM-Decoding bias and threshold}. \textit{We observe that both the mean accepted tokens (MAT) and the speedup ratio increase with $l_{\rm bias}$ and $l_{\rm threshold}$ before they equal 5}. When the value of both parameters exceeds 5, these indicators begin to decline.

Additionally, we investigated how the draft length utilized by SAM-Decoding affects inference speed. Figure \ref{fig: SAM-Decoding draft size} illustrates the throughput of SAM-Decoding[T] at varying draft sizes. \textit{As the draft size increases, there is a positive trend in throughput until the draft size equals 40}. When the draft size exceeds 40, there is an observable decline in performance metrics, which becomes more significant as the draft size reaches 70. This phenomenon can be attributed to the fact that, for draft sizes below the average acceptance length, increasing the draft size reduces the number of rounds for generation, thereby enhancing efficiency. In contrast, once the draft size surpasses this threshold, further increases do not yield additional benefits and strain GPU capacity, thus slowing inference speed.




\begin{table}[t]
\newcolumntype{?}{!{\vrule width 1pt}}
\newcolumntype{C}{>{\centering\arraybackslash}p{2em}}
\centering
\renewcommand\arraystretch{1.5}
\resizebox{0.9\linewidth}{!}{
\begin{tabular}{l|ccc}
\hline
\multirow{2}{*}{Method}       & \multicolumn{3}{c}{Spec-Bench}  \\ \cline{2-4} 
                              & \#MAT & Tokens/s & Speedup      \\ \hline
PLD                           & 1.75  & 59.02    & 1.56$\times$ \\
SAM-Decoding                  & 2.30  & 69.37    & 1.84$\times$ \\
\quad w/o Static SAM          & 1.85  & 61.93    & 1.64$\times$ \\
\quad w/o Dynamic SAM         & 1.63  & 50.37    & 1.33$\times$ \\ \hline
\end{tabular}
}
\caption{The impact of different draft generation modules on inference speed.}
\label{table: draft modules}
\end{table}
Finally, we investigated the impact of different modules within SAM-Decoding on inference speed. SAM-Decoding comprises two draft generation modules: the static suffix automaton and the dynamic suffix automaton. We measured the inference speed of SAM-Decoding after removing each of these two modules individually. The results are presented in Table \ref{table: draft modules}. From the experimental results, it is clear that each module contributes to the acceleration of the decoding process. Notably, the dynamic suffix automaton has a significantly greater impact compared to the static suffix automaton. \textit{This suggests that, in many cases, generating drafts from the dynamic context is more effective than retrieving drafts from a pre-existing text corpus}. For more ablation experiment results, please refer to Appendix \ref{section: Additional Ablation Experiments}.


\section{Related Work}


\vpara{Speculative Decoding.} Speculative decoding is an approach that can significantly speed up large language models (LLMs) without compromising the quality of their outputs. The majority of speculative decoding techniques rely on smaller neural networks to create drafts during the inference process. These techniques are referred to as model-based speculative decoding methods. Early implementations of model-based speculative decoding, such as those Speculative Decoding~\cite{spec-decoding-leviathan-2023}, primarily focused on generating draft sequences using pre-existing, smaller-scale LLMs. Subsequently, advancements like Medusa~\cite{medusa-cai-2024}, SpecInfer~\cite{specinfer-miao-2024} and EAGLE~\cite{eagle-li-2024, eagle2-li-2024} introduced tree-based speculative methods and began the development of draft models tailored for speculative decoding. 



In contrast to model-based methods, certain approaches focus on generating drafts through retrieval, utilizing $n$-gram matching, which we refer to the retrieval-based method. Notable among these are Lookahead Decoding \cite{lookahead-fu-2024}, PIA\cite{pia-lookahead-zhao-2024}, PLD \cite{pld-saxena-2023} and REST \cite{rest-he-2024}. Token Recycling~\cite{token-recycle-luo-2024}, on the other hand, utilizes the previously generated token distribution to generate drafts, becoming a model-free method different from the retrieval-based method.


Additionally, beyond the aforementioned methods, research also conducted on speculative decoding that relies either on the model itself \cite{cllms-kou-2024} or on sub-models within the larger architecture \cite{layerskip-elhoushi-2024}.

\vpara{Efficient LLM Architecture.} There is also work to improve the model's inference speed from the perspective of model structure. This part of the work includes model distillation, quantization and pruning. Model distillation~\cite{minitron-sreenivas-2024, compact-llm-muralidharan-2024} distills the knowledge of a large model into a small model thereby speeding up inference while maintaining the model's performance. Quantization~\cite{gptq-frantar-2022, smoothquant-xiao-2023, awq-lin-2024, spinquant-liu-2024, quarot-ashkboos-2024} reduces the number of bits required to store parameters and reduces the data transmission time from HBM to on-chip memory during inference. Pruning~\cite{sparsegpt-frantar-2023, slicegpt-ashkboos-2024, shortgpt-men-2024, llm-streamline-chen-2024, sp3-hu-2024, wanda-sun-2024, RIA-zhang-2024} is used to remove unimportant parameters in the model. For structured pruning, it can be combined with model distillation to train efficient small models, while semi-structured pruning can reduce the model's memory access and computing overhead and improve the inference speed by combining special hardware.

\section{Conclusion}

In this work, we propose \model, an speculative decoding method via suffix automatons constructed from both generated text and text corpus. \smodel can efficiently retrieve drafts from retrieval sources, thereby accelerating inference. \smodel is also designed to seamlessly integrate with existing SD methods. Consequently, in scenarios where retrieval is not feasible, \model can adaptively switch to alternative methods for draft generation. Experimental results demonstrate that \model outperform retrieval-based SD baselines. Meanwhile, when combined with state-of-the-art techniques, \model can significantly enhance their performance in Multi-turn Conversation, Summarization, Retrieval-augmented Generation, and Context Q\&A tasks.


\section{Limitation}

On the one hand, as a retrieval-based speculative decoding method, the performance of SAM-Decoding depends on the task type as well as the quality of the retrieval source. Currently, we have collected a text corpus based on the vicuna-7b generated results on Stanford-alpaca, GSM8k and python-instruct-18k. However, this corpus is still not diverse enough, and also the text in it may deviate from the text generated by other LLMs, which limits the performance of SAM-Decoding. Therefore, in the future we need to collect more specialized and diverse corpus for different types of tasks.

On the other hand, when combining SAM-Decoding with other types of methods, we use a very heuristic approach, i.e., we choose different methods depending on the match length. This does not fully utilize the exact match lengths provided by the suffix automaton, so subsequently we will try to train classifier to select different decoding methods at each generate round.

Finally, the performance of retrieval-based methods is highly correlated with the usage scenarios, and the existing datasets do not well reflect the performance of retrieval-based methods in real usage, so in the future we also need to construct datasets that are more compatible with real scenarios to evaluate the performance of retrieval-based methods.

\bibliography{custom}

\appendix

\section{Suffix Automaton}

\subsection{Construction Process of Suffix Automaton}
\label{appendix: Construction Process of Suffix Automaton}

Algorithm \ref{algo: Construction Process of Suffix Automaton} introduces the construction (Build-SAM) and expansion process (Expand) of Suffix Automaton, where the INIT\_SAM function will create a suffix automaton that only contains the root node. For the root node, the \textbf{link} attribute value is \(-1\), the \textbf{next} attribute value is empty, the length attribute value is \(0\), and the min\_endpos attribute value is \(0\). Meanwhile, Algorithm~\ref{algo: Initialize top-k successor} shows the construction process of the top-k successors for each node of static suffix automaton. Each node in the algorithm involves a new variable, ``freq'', which represents the frequency of occurrence of the corresponding substring for each node, and can be initialized at the time of constructing the suffix automaton, i.e., ``freq'' is initialized to 1 for nodes generated by expansion, and ``freq'' is initialized to 0 for nodes generated based on cloning.

\begin{algorithm}[t]
    \caption{Construction Process of Suffix Automaton}
    \label{algo: Construction Process of Suffix Automaton}
\begin{algorithmic}
    \FUNCTION{Expand-State}
       \STATE {\bfseries Input:} suffix automaton $S$, link $l$, next $n$, length $len$, position $p$
       \STATE $s = S.\text{expand\_state()}$
       \STATE $s.\text{link} = l$
       \STATE $s.\text{next} = n$
       \STATE $s.\text{length} = len$
       \STATE $s.\text{min\_endpos} = p$
       \STATE {\bfseries Output:} new state $s$
    \ENDFUNCTION
    \FUNCTION{Expand}
        \STATE {\bfseries Input:} suffix automaton $S$, token $t$
        \STATE $S.\text{max\_length} = S.\text{max\_length} + 1$
        \STATE $l = S.\text{max\_length}$
        \STATE $c = \text{Expand-State}(S, -1, \{\}, l, l)$
        \STATE $p = S.\text{last}$
        \WHILE{$p \neq -1 $ \AND $t \notin p\text{.next}$}
            \STATE $p.\text{next[t]} = c$
            \STATE $p = p.\text{link}$
        \ENDWHILE
        \IF{$p = \textbf{None}$}
            \STATE $c.\text{link} = S.\text{root}$
        \ELSE
            \STATE $q = p.\text{next[t]}$
            \IF{p.\text{length} + 1 = q.\text{length}}
                \STATE $c.\text{link} = q$
            \ELSE
                \STATE $cl = \text{Expand-State}(S, , -1, \{\}, -1, -1)$
                \STATE $cl.\text{link} = q.\text{link}$
                \STATE $cl.\text{next} = q.\text{next}$
                \STATE $cl.\text{length} = p.\text{length} + 1$
                \STATE $cl.\text{min\_endpos} = q.\text{min\_endpos}$
                \WHILE{$p \neq \textbf{None}$ \AND $p.\text{next[t]} = q$}
                    \STATE $p.\text{next[t]} = cl$
                    \STATE $p = p.\text{link}$
                \ENDWHILE
                \STATE $q.\text{link} = c.\text{link} = cl$
            \ENDIF
        \ENDIF
        \STATE $S.\text{last} = c$
    \ENDFUNCTION
    \FUNCTION{Build-SAM}
        \STATE {\bfseries Input:} token sequence $s$
        \STATE $S = \text{INIT\_SAM}()$
        \FOR{$t$ {\bfseries in} $s$}
            \STATE $\text{Expand}(S, t)$
        \ENDFOR
       \STATE {\bfseries Output:} suffix automaton $S$
    \ENDFUNCTION
\end{algorithmic}
\end{algorithm}

\begin{algorithm}[t]
    \caption{Construction Process of Top-k Successors and Transition Probabilities}
    \label{algo: Initialize top-k successor}
\begin{algorithmic}
    \FUNCTION{dfs}
       \STATE {\bfseries Input:} state $s$
       \FOR{$t_n, s_n \in s.\text{next}$}
            \STATE $\text{dfs}(s_n)$
            \STATE $s.\text{freq} = s.\text{freq} + s_n.\text{freq}$
       \ENDFOR
       \STATE $s.\text{topk\_succs} = \text{TopK}_\text{freq}(s.\text{next})$
       \STATE $s.\text{topk\_prob} = []$
       \FOR{$t_n, s_n \in s.\text{topk\_succ}$}
            \STATE $s.\text{topk\_prob}.\text{append}(s_n.\text{freq} / s.\text{freq})$
       \ENDFOR
    \ENDFUNCTION
    \FUNCTION{Init\_topk}
       \STATE {\bfseries Input:} suffix automaton $S$
       \STATE $\text{dfs}(S.\text{root})$
    \ENDFUNCTION
\end{algorithmic}
\end{algorithm}

\subsection{Drafting via Prim's Algorithm}
\label{appendix: Drafting via Prim's Algorithm}

Algorithm \ref{algo: Drafting via Prim's Algorithm} introduces a drafting process based on Prim's algorithm to find a maximum spanning tree. For static suffix automata, we can offline maintain the frequency of occurrence of the corresponding substring for each node. Therefore, based on the recorded frequency for each node in the automaton, we can calculate the top-k successors and corresponding transition probabilities, where the transition probability is calculated by dividing the frequency of occurrence of the target state by the frequency of occurrence of the current state. 

\begin{algorithm}[t]
    \caption{Drafting via Prim's Algorithm}
    \label{algo: Drafting via Prim's Algorithm}
\begin{algorithmic}
    \FUNCTION{Prim}
       \STATE {\bfseries Input:} suffix automaton $S$, state $s$, start token $t$
       \STATE $q = \text{PriorityQueue}()$
       \STATE $q.\text{push}(\{1.0, s, t\})$
       \STATE $d = []$
       \WHILE{$q.\text{size}() > 0$ \\ \AND $d.\text{size}() \neq \text{MAX\_SIZE}$}
            \STATE $p,\, s,\, t = q.\text{top}()$
            \STATE $q.\text{pop}()$
            \STATE $d.\text{append}(t)$
            \FOR{$(t_n,s_n,p_n)$ {\bfseries in} \\ $\text{zip}(s.\text{topk\_succ}, s.\text{topk\_prob})$}
                \STATE $p_{\rm new} = p * p_n$
                \STATE $s_{\rm new} = s_n$
                \STATE $t_{\rm new} = t_n$
                \STATE $q.\text{push}({p_{\rm new}, s_{\rm new}, t_{new}})$
            \ENDFOR
       \ENDWHILE
       \STATE {\bfseries Output:} draft tree $d$
    \ENDFUNCTION

\end{algorithmic}
\end{algorithm}

\subsection{Time Complexity of State Transfer}
\label{appendix: Time Complexity of Suffix Automaton}

In this section, we introduce the time complexity of state transfer of suffix automaton. Consider a suffix automaton \( S \) with initial state \( s_0 \), which corresponds to the root node of the automaton (representing the empty string). Suppose that state \( s_0 \) undergoes transitions through a sequence of \( L \) tokens \( x = (x_1, x_2, \ldots, x_L) \):

\[
s_i = \text{Transfer}(S, x_i, s_{i-1}), \quad i \in \{1, 2, \ldots, L\}.
\]

We aim to demonstrate that the average time complexity of each state transition is \( O(1) \), while the worst-case time complexity is \( O(L) \).

First, let us define the matching length associated with state \( s_i \) as \( l_i \). Given that each state transition can increase the length of the match by at most 1, it follows that \( 0 \le l_i \le i \). Next, we introduce the concept of energy \( \phi \) for each state \( s_i \), defined as \( \phi(s_i) = l_i \). Let \( c_i \) represent the time cost of the transition of the \( i \)-th state. We then define the amortized cost \( \hat{c_i} \) as:

\[
\hat{c_i} = c_i + \phi(s_i) - \phi(s_{i-1}).
\]

We can now express the total amortized cost over all transitions as:

\begin{align*}
\sum_{i=1}^{L} \hat{c_i} & = \sum_{i=1}^{L} (c_i + \phi(s_i) - \phi(s_{i-1})) \\
                         & = \sum_{i=1}^{L} c_i + \phi(s_L) - \phi(s_0).    
\end{align*}

Since \( \phi(s_i) \ge 0 \) and \( \phi(s_0) = 0 \), it follows that:

\[
\sum_{i=1}^{L} \hat{c_i} \ge \sum_{i=1}^{L} c_i.
\]

Next, we analyze the upper bound of \( \hat{c_i} \). Each state transition involves moving through the \textbf{link} edge zero or more times, followed by a move through the \textbf{next} edge. Transitioning through the \textbf{link} edge incurs a cost of 1 but decreases the potential by at least 1. Conversely, transitioning through the \textbf{ next} edge incurs a cost of 1 and increases the potential by 1. Consequently, the amortized cost \( \hat{c_i} \) is bounded above by 2, leading to:

\[
\sum_{i=1}^{L} \hat{c_i} \le 2L.
\]

Thus, the average time complexity of state transitions is:

\[
\frac{\sum_{i=1}^{L} c_i}{L} \le \frac{2L}{L} = 2,
\]

\noindent which is \( O(1) \). In the worst case, a single operation may require up to \( l_i \) transitions through the \textbf{link} edge, followed by one transition through the \textbf{next} edge, resulting in a worst-case time complexity of \( O(L) \).

\begin{table*}[t]
\newcolumntype{?}{!{\vrule width 1pt}}
\newcolumntype{C}{>{\centering\arraybackslash}p{2em}}
\centering
\renewcommand\arraystretch{1.5}
\resizebox{\linewidth}{!}{
\begin{tabular}{llccccccccc}
\toprule
Model &  Method & \makecell{MT} & Trans & Sum & QA & Math & RAG & \#MAT & Tokens/s & Speedup \\ \hline
\multirow{6}{*}{Llama3-8B}
 & PLD       
 &  1.30$\times$ & 1.12$\times$  & 1.41$\times$ & 1.03$\times$  & 1.30$\times$  & 1.53$\times$ & 1.39  & 44.26 & 1.28$\times$ \\
 & \textbf{SAM-Decoding}
 & 1.59$\times$  & 1.35$\times$  & 1.50$\times$ & 1.35$\times$  & 1.54$\times$  & 1.75$\times$ & 1.72  & 52.35 & 1.51$\times$   \\ 
 & Token Recycling 
 & 1.92$\times$  & 1.88$\times$  & 1.85$\times$ & 1.75$\times$  & 2.24$\times$  & 1.82$\times$ & 2.76  & 66.42 & 1.91$\times$   \\ 
 & \textbf{SAM-Decoding{[}T{]}}
 & 2.09$\times$  & 1.93$\times$  & 2.04$\times$ & 1.82$\times$  & 2.32$\times$  & 2.12$\times$ & 2.63  & 71.73 & 2.05$\times$   \\ 
 & EAGLE-2  
 & 2.08$\times$   & 1.95$\times$   & 1.85$\times$   & 1.80$\times$   & 2.31$\times$  & 1.87$\times$     & 3.90   & 68.69 & 1.98$\times$   \\
 & \textbf{SAM-Decoding{[}E2{]}}
 & 2.36$\times$   & 1.96$\times$   & 1.98$\times$   & 1.79$\times$   & 2.32$\times$  & 2.11$\times$     & 3.92   & 72.47 & 2.08$\times$  \\ \hline
\end{tabular}
}
\caption{Speedup of SAM-Decoding compared to the baselines on Spec-Bench.}
\label{table: Spec-Bench-Llama3}
\end{table*}

\begin{table*}[t]
\newcolumntype{?}{!{\vrule width 1pt}}
\newcolumntype{C}{>{\centering\arraybackslash}p{2em}}
\centering
\renewcommand\arraystretch{1.5}
\resizebox{0.8\linewidth}{!}{
\begin{tabular}{ll|ccc|ccc}
\hline
\multirow{2}{*}{Model}          & \multirow{2}{*}{Method} & \multicolumn{3}{c|}{HumanEval}  & \multicolumn{3}{c}{HAGRID}      \\ \cline{3-8} 
                                &                         & \#MAT & Tokens/s & Speedup      & \#MAT & Tokens/s & Speedup      \\ \hline
\multirow{6}{*}{Llama3-8B}      & PLD                     & 1.30  & 42.39    & 1.18$\times$ & 1.50  & 45.15    & 1.56$\times$ \\
                                & \textbf{SAM-Decoding}     
                                                          & 2.06  & 64.38    & 1.79$\times$ & 1.88  & 58.40    & 2.02$\times$ \\
                                & Token Recycling         & 2.93  & 71.49    & 1.99$\times$ & 2.84  & 62.77    & 2.17$\times$ \\
                                & \textbf{SAM-Decoding{[}T{]}}     
                                                          & 2.77  & 78.04    & 2.16$\times$ & 2.70  & 66.76    & 2.30$\times$ \\
                                & EAGLE-2                  & 4.74  & 85.58    & 2.37$\times$ & 3.97  & 63.30    & 2.18$\times$ \\
                                & \textbf{SAM-Decoding{[}E2{]}}    
                                                          & 4.76  & 91.50    & 2.54$\times$ & 3.93  & 67.94    & 2.35$\times$ \\ \hline
\end{tabular}
}
\caption{Speedup of SAM-Decoding compared to the baselines on HumanEval and HAGRID.}
\label{table: humaneval HAGRID Llama3}
\end{table*}

\begin{table*}[t]
\newcolumntype{?}{!{\vrule width 1pt}}
\newcolumntype{C}{>{\centering\arraybackslash}p{2em}}
\centering
\renewcommand\arraystretch{1.5}
\resizebox{\linewidth}{!}{
\begin{tabular}{llccccccccc}
\toprule
Model &  Method & \makecell{MT} & Trans & Sum & QA & Math & RAG & \#MAT & Tokens/s & Overall \\ \hline
\multirow{6}{*}{Vicuna-13B}
 & PLD       
 &  1.61$\times$  & 1.10$\times$ & 2.36$\times$ & 1.11$\times$ & 1.69$\times$ & 1.80$\times$ & 1.66 & 33.89   & 1.59$\times$ \\
 & \textbf{SAM-Decoding}
 & 2.08$\times$  & 1.26$\times$  & 2.23$\times$  & 1.53$\times$  & 2.09$\times$  & 1.89$\times$ & 2.19  & 39.24 & 1.84$\times$   \\ 
 & Token Recycling 
 & 2.03$\times$  & 1.84$\times$ & 2.07$\times$ & 1.83$\times$  & 2.42$\times$  & 1.84$\times$  & 2.81  & 42.74 & 2.01$\times$   \\ 
 & \textbf{SAM-Decoding{[}T{]}}
 & 2.36$\times$  & 1.80$\times$  & 2.63$\times$  & 1.83$\times$  & 2.49$\times$  & 2.22$\times$ & 2.91  & 47.27 & 2.22$\times$   \\ 
 & EAGLE-2  
 & 3.10$\times$   & 2.15$\times$   & 2.58$\times$   & 2.38$\times$   & 3.19$\times$  & 2.33$\times$     & 4.42   & 56.06 & 2.63$\times$   \\
 & \textbf{SAM-Decoding{[}E2{]}}
 & 3.27$\times$   & 2.12$\times$   & 2.89$\times$   & 2.34$\times$   & 3.12$\times$  & 2.54$\times$     & 4.51   & 57.88 & 2.72$\times$  \\ \hline
\end{tabular}
}
\caption{Speedup of SAM-Decoding compared to the baselines on Spec-Bench.}
\label{table: Spec-Bench-Vicuna-13B}
\end{table*}

\begin{table*}[t]
\newcolumntype{?}{!{\vrule width 1pt}}
\newcolumntype{C}{>{\centering\arraybackslash}p{2em}}
\centering
\renewcommand\arraystretch{1.5}
\resizebox{0.8\linewidth}{!}{
\begin{tabular}{ll|ccc|ccc}
\hline
\multirow{2}{*}{Model}          & \multirow{2}{*}{Method} & \multicolumn{3}{c|}{HumanEval}  & \multicolumn{3}{c}{HAGRID}      \\ \cline{3-8} 
                                &                         & \#MAT & Tokens/s & Speedup      & \#MAT & Tokens/s & Speedup      \\ \hline
\multirow{6}{*}{Vicuna-13B}     & PLD                     & 1.54  & 32.06    & 1.44$\times$ & 1.90  & 43.38    & 2.15$\times$ \\
                                & \textbf{SAM-Decoding}     
                                                          & 2.42  & 48.92    & 2.20$\times$ & 2.21  & 41.93    & 2.08$\times$ \\
                                & Token Recycling         & 2.79  & 46.03    & 2.07$\times$ & 2.90  & 40.97    & 2.03$\times$ \\
                                & \textbf{SAM-Decoding{[}T{]}}     
                                                          & 2.79  & 50.87    & 2.28$\times$ & 2.99  & 48.33    & 2.40$\times$ \\
                                & EAGLE-2                  & 5.15  & 77.85    & 3.49$\times$ & 4.24  & 52.28    & 2.59$\times$ \\
                                & \textbf{SAM-Decoding{[}E2{]}}    
                                                          & 5.12  & 78.96    & 3.54$\times$ & 4.41  & 56.17    & 2.78$\times$ \\ \hline
\end{tabular}
}
\caption{Speedup of SAM-Decoding compared to the baselines on HumanEval and HAGRID.}
\label{table: humaneval HAGRID Vicuna-13B}
\end{table*}

\begin{table*}[t]
\newcolumntype{?}{!{\vrule width 1pt}}
\newcolumntype{C}{>{\centering\arraybackslash}p{2em}}
\centering
\renewcommand\arraystretch{1.5}
\resizebox{\linewidth}{!}{
\begin{tabular}{llccccccccc}
\toprule
Model &  Method & \makecell{MT} & Trans & Sum & QA & Math & RAG & \#MAT & Tokens/s & Overall \\ \hline
\multirow{6}{*}{Vicuna-33B}
 & PLD       
 &  1.50$\times$  & 1.07$\times$ & 2.06$\times$ & 1.09$\times$ & 1.59$\times$ & 1.51$\times$ & 1.65 & 13.33   & 1.46$\times$ \\
 & \textbf{SAM-Decoding}
 & 1.91$\times$  & 1.25$\times$ & 1.98$\times$ & 1.48$\times$  & 1.83$\times$  & 1.66$\times$  & 1.97  & 15.35 & 1.68$\times$   \\ 
 & Token Recycling 
 & 2.10$\times$  & 1.84$\times$ & 2.19$\times$ & 1.88$\times$  & 2.42$\times$  & 1.92$\times$  & 2.70  & 18.80 & 2.06$\times$   \\ 
 & \textbf{SAM-Decoding{[}T{]}}
 & 2.31$\times$  & 1.79$\times$ & 2.53$\times$ & 1.90$\times$  & 2.48$\times$  & 2.06$\times$  & 2.68  & 19.87 & 2.18$\times$   \\ 
 & EAGLE-2  
 & 3.29$\times$   & 2.31$\times$   & 2.73$\times$   & 2.51$\times$   & 3.65$\times$  & 2.46$\times$     & 4.06   & 25.86 & 2.83$\times$   \\
 & \textbf{SAM-Decoding{[}E2{]}}
 & 3.40$\times$   & 2.25$\times$   & 2.93$\times$   & 2.43$\times$   & 3.45$\times$  & 2.54$\times$     & 4.08   & 25.91 & 2.84$\times$  \\ \hline
\end{tabular}
}
\caption{Speedup of SAM-Decoding compared to the baselines on Spec-Bench.}
\label{table: Spec-Bench-Vicuna-33B}
\end{table*}

\begin{table*}[t]
\newcolumntype{?}{!{\vrule width 1pt}}
\newcolumntype{C}{>{\centering\arraybackslash}p{2em}}
\centering
\renewcommand\arraystretch{1.5}
\resizebox{0.8\linewidth}{!}{
\begin{tabular}{ll|ccc|ccc}
\hline
\multirow{2}{*}{Model}          & \multirow{2}{*}{Method} & \multicolumn{3}{c|}{HumanEval}  & \multicolumn{3}{c}{HAGRID}      \\ \cline{3-8} 
                                &                         & \#MAT & Tokens/s & Speedup      & \#MAT & Tokens/s & Speedup      \\ \hline
\multirow{6}{*}{Vicuna-33B}     & PLD                     & 1.58  & 14.18    & $1.51\times$ & 1.55  & 15.74    & $1.80\times$ \\
                                & \textbf{SAM-Decoding}     
                                                          & 2.05  & 19.08    & $2.03\times$ & 1.90  & 16.15    & $1.85\times$ \\
                                & Token Recycling         & 2.64  & 19.64    & $2.09\times$ & 2.71  & 18.29    & $2.09\times$ \\
                                & \textbf{SAM-Decoding{[}T{]}}     
                                                          & 2.73  & 22.44    & $2.39\times$ & 2.60  & 19.74    & $2.26\times$ \\
                                & EAGLE-2                  & 3.53  & 28.18    & $3.00\times$ & 3.84  & 24.28    & $2.78\times$ \\
                                & \textbf{SAM-Decoding{[}E2{]}}    
                                                          & 3.61  & 29.56    & $3.14\times$ & 3.82  & 25.08    & $2.87\times$ \\ \hline
\end{tabular}
}
\caption{Speedup of SAM-Decoding compared to the baselines on HumanEval and HAGRID.}
\label{table: humaneval HAGRID Vicuna-33B}
\end{table*}

\begin{table*}[t]
\newcolumntype{?}{!{\vrule width 1pt}}
\newcolumntype{C}{>{\centering\arraybackslash}p{2em}}
\centering
\renewcommand\arraystretch{1.5}
\resizebox{\linewidth}{!}{
\begin{tabular}{llccccccccc}
\toprule
Model &  Method & \makecell{MT} & Trans & Sum & QA & Math & RAG & \#MAT & Tokens/s & Overall \\ \hline
\multirow{4}{*}{Vicuna-7B}
 & Token Recycling 
 & 2.08$\times$  & 1.76$\times$ & 1.97$\times$ & 1.85$\times$  & 2.35$\times$  & 1.76$\times$  & 2.82  & 98.39 & 1.96$\times$   \\ 
 & \textbf{SAM-Decoding{[}T{]}}
 & 2.62$\times$  & 1.82$\times$  & 2.92$\times$  & 2.09$\times$  & 2.60$\times$  & 2.21$\times$ & 3.02  & 119.21 & 2.38$\times$   \\ 
 & EAGLE-2  
 & 2.66$\times$   & 1.76$\times$   & 2.18$\times$   & 2.03$\times$   & 2.63$\times$  & 1.97$\times$     & 4.34   & 110.56 & 2.21$\times$   \\
 & \textbf{SAM-Decoding{[}E2{]}}
 & 3.19$\times$   & 1.97$\times$   & 2.86$\times$   & 2.28$\times$   & 2.84$\times$  & 2.32$\times$     & 4.52   & 129.36 & 2.58$\times$  \\ \hline
\end{tabular}
}
\caption{Speedup of SAM-Decoding on A800 GPU compared to the baselines on Spec-Bench.}
\label{table: Spec-Bench-Vicuna-7B-A800}
\end{table*}


\section{Additional Experiment Results}
\label{section: Additional Experiment Results}

In this section, we present the results of the experiment on Llama3-8B-instruct, Vicuna-13B-v1.3 and Vicuna-33B-v1.3. 

Tables \ref{table: Spec-Bench-Llama3} and \ref{table: humaneval HAGRID Llama3} present the speedup ratios of \smodel compared to baseline methods across the Spec-Bench, HumanEval, and HAGRID datasets, utilizing the Llama3-8B-instruct model. It can be seen that the inference speed of SAM-Decoding outperforms the strongest retrieval-based baseline PLD on all tasks. Meanwhile, \smodel, when paired with Token Recycling (SAM-Decoding[T]), brings speedups on all tasks. Specifically, \smodel enhances the speedup ratio of Token Recycling from 1.92$\times$, 1.85$\times$, and 1.82$\times$ to 2.09$\times$, 2.04$\times$, and 2.12$\times$ for Multi-turn Conversation, Summarization, and Retrieval-Augmented Generation tasks, respectively. This improvement raises the overall speedup ratio of token recycling in the Spec-Bench dataset from 1.91$\times$ to 2.05$\times$. On the HumanEval and HAGRID datasets, \smodel increases the speedup ratio of Token Recycling from 1.99$\times$ and 2.17$\times$ to 2.16$\times$ and 2.30$\times$, respectively. Furthermore, \smodel also amplifies the performance gains of EAGLE-2 in Multi-turn Conversation, Summarization, Retrieval-augmented Generation, Code Generation and Context Q\&A tasks. The speedup ratios were increased from 2.08$\times$, 1.85$\times$, 1.87$\times$, 2.37$\times$, and 2.18$\times$ to 2.36$\times$, 1.98$\times$, 2.11$\times$, 2.54$\times$ and 2.35$\times$ respectively.

Tables \ref{table: Spec-Bench-Vicuna-13B}, \ref{table: humaneval HAGRID Vicuna-13B}, \ref{table: Spec-Bench-Vicuna-33B} and \ref{table: humaneval HAGRID Vicuna-33B} present the speedup ratios of \smodel compared to baseline methods across the Spec-Bench, HumanEval, and HAGRID datasets, utilizing the Vicuna-13B-v1.3 and Vicuna-33B-v1.3. On both models, SAM-Decoding still has inference speed exceeding the retrieval-based baseline, while by combining Token Recycling and EAGLE-2 also further improves the inference speed of the model on the Multi-turn Conversation, Summarization, Retrieval-augmented Generation and Context Q\&A tasks.

\section{Additional Ablation Experiments}
\label{section: Additional Ablation Experiments}

\begin{figure}[t]
    \centering
    \includegraphics[width=0.9\linewidth]{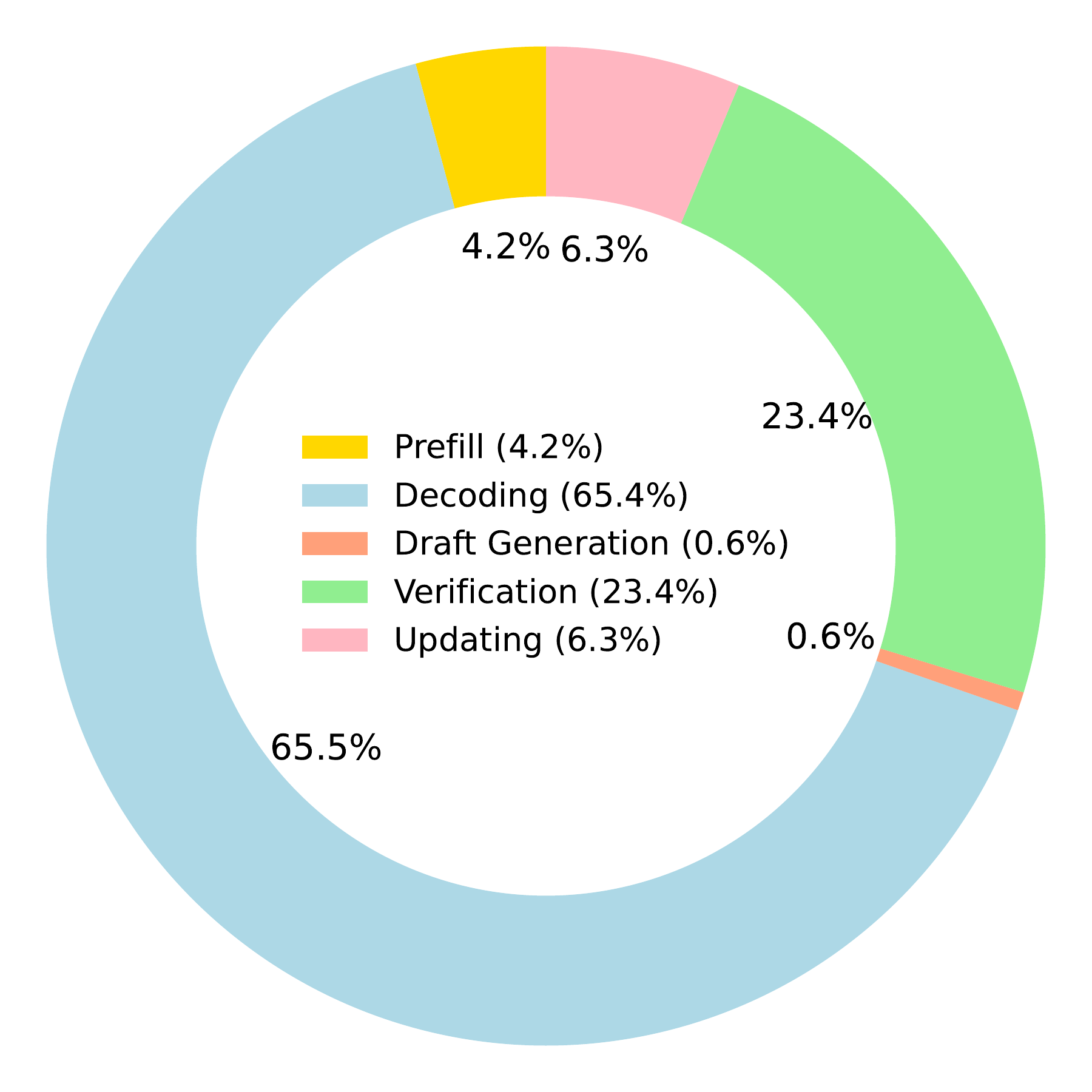}
    \caption{the percentage of inference time of different modules in \model.}
    \label{fig: Time Distribution}
\end{figure}

\begin{figure}[t]
    \centering
    \includegraphics[width=0.9\linewidth]{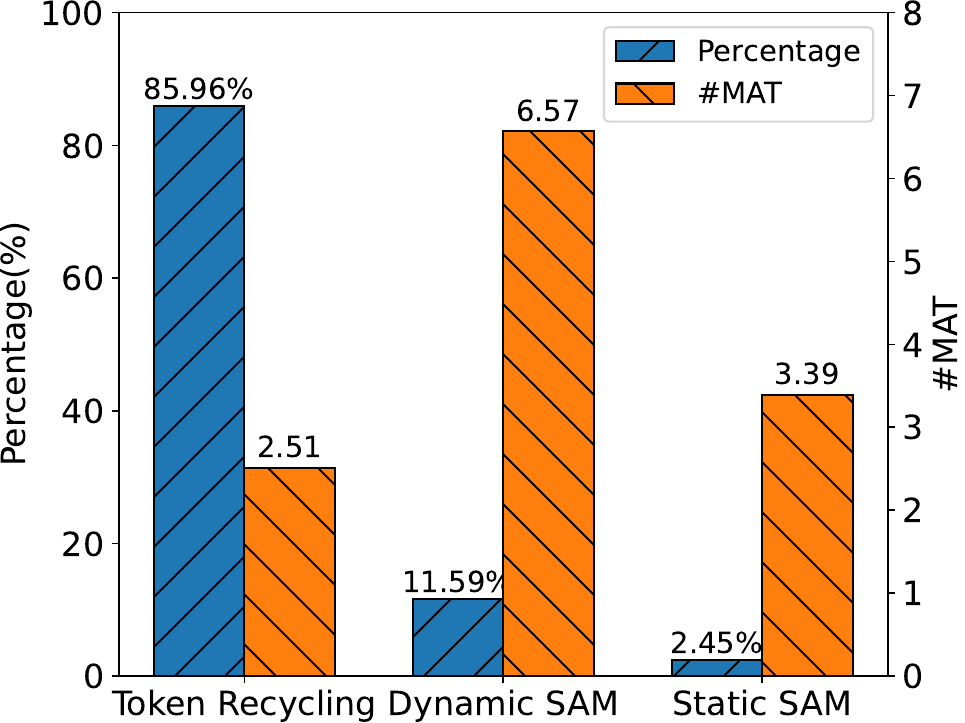}
    \caption{the percentage of usage and mean accept tokens of different draft modules.}
    \label{fig: Draft Distribution}
\end{figure}

In this section, we present additional ablation experiments, including the percentage of inference time of different modules in the decoding process of \model, and the percentage of drafts provided by different draft modules in \model. 

The inference process of \smodel is divided into five stages: prefill, draft generation, decoding, verification, and updating. During the prefill stage, the model processes the input prompt to establish an initial state. In the first draft generation stage, a draft is produced based on this initial state. The decoding stage involves the model further processing this draft. Next comes verification, where the correct parts of the draft are evaluated based on the information processed during the decoding stage. Finally, the update phase modifies the state of the model based on the valid parts of the draft. Figure \ref{fig: Time Distribution} illustrates the proportion of time each stage consumes within the SAM-Decoding[T] process based on Spec-Bench. As shown, the decoding stage takes up the largest portion of time, accounting for 65.4\% of the entire process. This is followed by the verification stage, which occupies 23.4\% of the total time. The updating stage requires 6.3\% of the time, whereas the draft generation stage contributes only 0.6\% to the overall duration. Additionally, the prefill stage comprises 4.2\% of the total processing time.

Figure \ref{fig: Draft Distribution} shows the usage frequency of different draft modules of SAM-Decoding[T] on Spec-Bench and the corresponding average draft accept length. It can be seen that in 85.96\% of the cases, due to insufficient matching length, we generate drafts based on the auxiliary method, corresponding to an average accept length of 2.51, while in the remaining 11.59\% and 2.45\% of the cases, the dynamic suffix automaton and static suffix automaton are used to generate drafts, corresponding to average accept lengths of 6.57 and 3.39, respectively.

Finally, Table \ref{table: Spec-Bench-Vicuna-7B-A800} shows the inference speed of different methods based on Vicuna-7B-v1.3 on NVIDIA A800 GPU. It can be seen that SAM-Decoding can still effectively combine Token Recycling and EAGLE-2 to achieve higher inference speed, which shows the effectiveness of our approach for different devices.

\end{document}